# Input Scheme for Hindi Using Phonetic Mapping

**Nisheeth Joshi and Iti Mathur**

**Apaji Institute, Banasthali University, Rajasthan**
Email: nisheeth.joshi@rediffmail.com, mathur_iti@rediffmail.com

## Abstract

*Written Communication on Computers requires knowledge of writing text for the desired language using Computer. Mostly people do not use any other language besides English. This creates a barrier. To resolve this issue we have developed a scheme to input text in Hindi using phonetic mapping scheme. Using this scheme we generate intermediate code strings and match them with pronunciations of input text. Our system show significant success over other input systems available.*

## 1. Introduction

Text input research has been going on since 1970s [1] and keyboard has been the most important device, as far as input is concerned and for a foreseeable future, is going to dominate the input system domain. Although we have alternatives like speech inputs and hand writing recognition systems, but their usage and performance has to undergo a lot of improvement. As of now, Text Input Systems rule the Input Domain. Hindi and most of the languages in South and South East Asian perspective have more characters then are available in English. So, trying to develop a text input system in these languages require a lot of processing and mapping. In this paper, we present an approach based on phonetic mapping where we try to input text for Hindi, using English. Here our work was two fold, firstly to develop a user friendly and efficient input system for Hindi based on phonetic mapping. Secondly, to evaluate the performance of the system developed as compared to other systems.

Hindi is the official language of India and enjoys the status of *Raj Bhasha*. It is the world's third most spoken language[2] (After Chinese and English). The language has been categorized under Indo-European: Indo-Iranian: Indo-Aryan family of languages[3]. It is also an official language of Republic of Fiji and Republic of Mauritius. It is the descendant of Sanskrit and Prakrit. In seventh century it started to emerge as a new language called Apabhramsha and by the end of tenth century it became stable and matured to a complete language.

In the eleventh century, during the invasion of Turks, the language was re-christened by them as Hindvi i.e. the language of the people of Hindukush valley for Hind. Due to the rule of the Turks on Hind, this language then came in contact with Persian (The language spoken by Turks) and through Persian, took some words of Arabic. This new fusion of language which was somewhat broken and mixed became the communication medium between the new inhabitants and the natives of Hind. Then, during eighteenth century, in the colonial rule of British Empire this language become the national language, because the British started to promote it as the standard language for official documents. During the next centaury the language was again renamed, but this time was given two names Hindi which was written in Devnagari Script which was taken from Sanskrit and Urdu which was written in Perso-Arabic Script.

As Hindi owe its origins from Sanskrit. It is an extremely logical and straight forward language. Perhaps this is the quality which makes it easier to learn. Pronunciation of Hindi is also very easy and is written in the same way as it is pronounced. Unlike English, this language does not have any capital letters and supports three forms of honorifics, making conversations either formal or familiar or intimate. Thus, the politeness expressed in English using words like 'Please', 'Thank you' etc. is already inherent in the language.

| | Short Vowels (हस्व स्वर) | | | Long Vowels (दीर्घ स्वर) | |
|---|---|---|---|---|---|
| **Vowels (स्वर)** | अ<br>a | इ<br>i | | आ<br>aa | ई<br>ii |
| | उ<br>u | ए<br>e | | ऊ<br>uu | ऐ<br>ai |
| | ओ<br>o | ऋ<br>ri | | औ<br>au | |
| **Diacritics** | ○<br>m | ○:<br>h | | | |
| **Consonants (व्यंजन)** | क<br>k | ख<br>kh | ग<br>g | घ<br>gh | ङ<br>nga |
| | च<br>c | छ<br>ch | ज<br>j | झ<br>jh | ञ<br>nja |
| | ट<br>ta | ठ<br>tha | ड<br>da | ढ<br>dha | ण<br>na |
| | त<br>t | थ<br>th | द<br>d | ध<br>dh | न<br>n |
| | प<br>p | फ<br>ph | ब<br>b | भ<br>bh | म<br>m |
| | य<br>y | र<br>r | ल<br>l | व<br>v | श<br>sh |
| | ष<br>ssh | स<br>s | ह<br>h | | |

Table 1: Hindi Character Set with Double Metaphone Phonetic Encoding

As shown in Table 1, Hindi language has 11 vowels also known as स्वर and 33 consonants also known as व्यांगन and 2 diacritics. Vowels can further be classified as short vowels and long vowels. Moreover vowels when used with consonants change their structure and become vowel symbols and are known as grapheme clusters, as shown in Table 2. A grapheme cluster is described as "what end users usually think of as character"[4]. As an exception, only one consonant 'र' can be transformed as a symbol and can take two forms as shown in the following table.

## 2. Review of Literature

We have witnessed great improvements in local language computing, since Government of India has put in crores of rupees into research in the area[1]. We can broadly classify all the research done in Input Method Technology (in Indian Perspective) into the following categories.

| Vowel Symbol → Consonant ↓ | अ | आ ा | इ ि | ई ी | उ ु | ऊ ू | ए े | ऐ ै | ओ ो | औ ौ | ऋ ृ | Consonant Before र | Consonant After र |
|---|---|---|---|---|---|---|---|---|---|---|---|---|---|
| क | क | का | कि | की | कु | कू | के | कै | को | कौ | कृ | क्र | र्क |
| च | च | चा | चि | ची | चु | चू | चे | चै | चो | चौ | चृ | च्र | र्च |
| ट | ट | टा | टि | टी | टु | टू | टे | टै | टो | टौ | टृ | ट्र | र्ट |
| र | र | रा | रि | री | रु | रू | रे | रै | रो | रौ | रृ |  | र्र |
| ल | ल | ला | लि | ली | लु | लू | ले | लै | लो | लौ | लृ | ल्र | र्ल |

Table 2: Example of Some Grapheme Clusters for Hindi

## 2.1 Direct Input Method

In this method an ASCII value is assigned to each Hindi alphabet/symbol. Central Institute of Indian Languages has also developed a coding scheme which is also based on the same lines and is named Indian Standard for Coded Information Interchange (ISCII). As shown in Figure 1(a), Users either have to input these codes or have to remember the mapping of each key on the keyboard to the corresponding character, making it very difficult to input text for new or first time users of Hindi on Computers. At present there are fonts like Mangal, Shruti etc., which are available in Unicode but they too suffer with the same drawbacks.

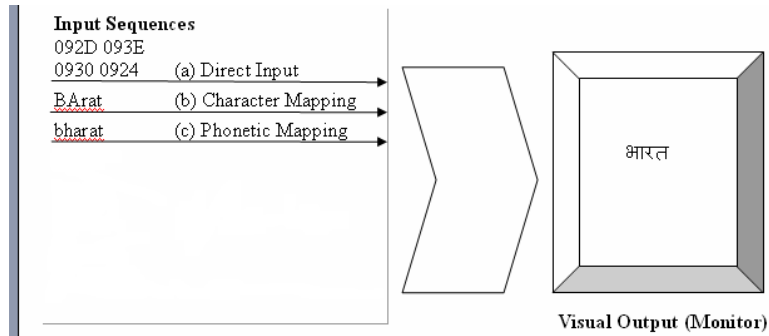

Figure1: Different Input Methods

## 2.2 Transliteration

Transliteration is a process of writing text of a language in some other language. Transliteration of English to other languages have been a popular research areas. Work on English to Hindi transliteration began in late 1980s, when an application named ITRANS[5] was developed to input text in Hindi using English. This scheme is also known as phonetic mapping, as shown in Figure 1(c). There is another form of transliteration where a character of English is mapped to a Hindi character, as shown in Figure 1(b). Since basis of ITRANS like systems is English spellings and pronunciations, it hits a bottleneck as Hindi has a series of characters which are phonetically same. A detailed discussion on these system can be referred in our previous work[6].

Although it takes care of ambiguity to some extent, it does not, at times give a intuitive sequence of inputs as sometimes the user has to write some text in fairly rigid manner. For example, If we wish to transliterate "jaipur rajasthan ki rajdhani hai". The phonetic transliteration of this text is "जैपुर रजस्थन कि रजधनि है". In order to get correct text i.e. "जयपुर राजस्थान की राजधानी है", we have to write "japura raajasthana kee raajadhaanee hai". As it is clearly visible form the example that by using this Input Method we panelize the beauty and elegance of the target language.

---
[1] http://tdil.mit.gov.in

# 3. Proposed System

In our phonetic mapping system, we use phonetic transliteration as the basis, but improved it by using a phonetic lexicon based system. The first step was to encode each word on the basis of lexicon. Due to Hindi's complex orthographic rules, this task was highly cumbersome. Using double Metaphone approach we resolved these issues. The entire process is explained as follows:

## 3.1 Creation of Lexicon Database and Character Table

As shown in table 1. we modified each character by associating an additional character to the vowels, as in 'u' for 'उ' and 'uu' for 'ऊ' and associated an additional h to consonants, as in 'c' for 'च' and 'ch' for 'छ'. This resolved the ambiguity of using character with similar phonemes. Once the entire character table was generated, our second work was to encode the sequences of characters as given by different people. We conducted this study because, many times while chatting or reading mails in Romanized Hindi, we encountered the variations in spellings of different individuals. So, it was a general assumption that different people tend to spell same words differently.

We conducted a study on 50 subjects, where we gave them a three page document in Hindi and asked them to convert the same in Romanized Hindi. They were given a half an hour time, so that a more spontaneous and intuitive spelling could be captured, as a little time was given to them to think for what would be the correct spelling of a word. Once this was done, we studied, how each Hindi character is written in English. Using this we generated a new table where we could map all the possible sequence of a Hindi characters. This allowed us to map many to many correspondence between characters of English and characters of Hindi. Some of the many to many mapped characters are shown in Figure 2.

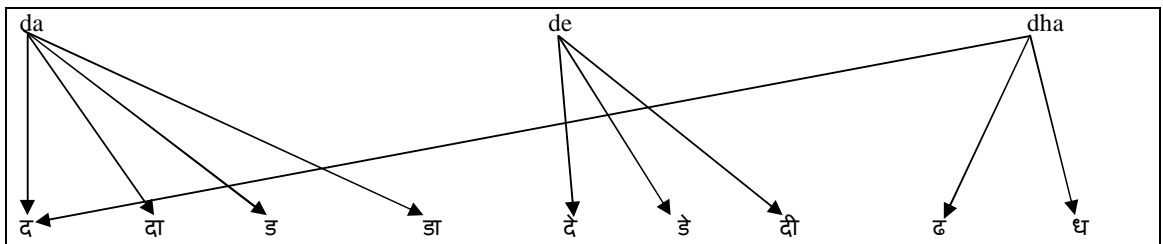
Figure 2: Many to Manny Correspondence between Hindi and Roman Alphabets

Once we have a table of all such correspondences, we created our lexicon database. We used the archives of Danik Bhaskar[2] (Hindi Daily Newspaper) for creating our lexicon database. We took the individual Hindi words and provided their English transliterations. We also calculated the frequencies of all these words. In all we created a database of 20,000 words.

## 3.2 Working

When the user enters the text, his input sequence is divided into characters using the character table and the right sequence of word, which can be matched in the database, is constructed internally. Once this is done the complete word or partial word is searched in the database. The matched words are ranked based on their frequencies. If there is only one word for the sequence then it is displayed in place of input sequence. If there are multiple matches then the top five matches (ones which have higher frequencies are displayed). If in case we do not get any word from the database then the characters are transcribed using direct mapping of character table. The entire algorithm is shown in figure 3.

---
[2] http://www.danikbhaskar.com

1. Generate phonetic codes for all Hindi words and create character table.
2. Input Hindi word using Roman (English) characters.
3. Generate phonetic code string of the input using character table.
4. IF input is phonetic code constructed through character table, then.
    a. If matches only one word in the lexicon, then
        i. Convert input to that Hindi word
    b. If matches to multiple words in the lexicon, then
        i. Produce suggestions of all relevant Hindi words and let the user select
5. if phonetic code does not match, then
    a. Use direct mapping of character table and provide the output

Figure 3: Algorithm for Hindi Input

## 4. Evaluation

We evaluated our system for user friendliness and efficiency. User Friendliness was calculated on ease of remembrance of words.

We asked a group of 60 people to write English equivalent of Hindi characters. The users were given a two column list, where in one column there was a Hindi character and in the other the user had to fill in English alphabet(s), based on their knowledge. Then we provided about an hour training to the users and gave sample exercises for Direct Input and Transliteration. Next, we gave two column list for both these schemes. We calculated edit distance based on each given encoding scheme, we calculated the given edit distance the user's transliteration and the scheme's transliteration.

The results are presented in Table 3. As shown, our predictive input system performed better then the two input schemes.

| Mapping Scheme | Average Edit Distance Per Word |
|---|---|
| Direct Input | 0.90 |
| Transliteration | 0.59 |
| Our Work | 0.34 |

Table 3. Edit Distance Experiment

Here average edit distance was calculated based on the following formulae:

1. $\text{Avg\_edit\_dist}(char) = \dfrac{1}{\#Subs} \sum_{subs=1}^{\#Subs} edit\_dist(inp\_seq(char), proposed(subs, char))$

2. $\text{Average} = \sum_{Char=1}^{\#Char} freq(char) \times avg\_edit\_dist(char)$

We calculated average edit distance between input key sequence and proposed text of each character. The average edit distance was calculated using the above equations. The results show that there is big difference between our system and Direct Input. Our average edit distance was far lesser then this input system. On the other hand difference between Transliteration and our system is very small. This is because people onto whom these tests were conducted tried to generate Romanized Hindi word which resembled English words i.e. they tried to avoid long vowel combinations like 'ii', 'uu', 'aa'.

## 5. Conclusion and Future Work

We have shown a methodology of taking input from a user who is not required to learn any input sequence. The use can input text as they deemed it correct. The system would understand the sequence and will try to predict the desired input by the user. We also experimentally proved that this method is user-friendly and efficient. This method can further be extended to other languages where there is no one to one correspondence with the roman key sequences.

As future directions for our work, we look forward to improve our system in many areas. First, we need to have a morpho-syntactic dictionary instead of the one that we are currently using. The dictionary then would take only the root word and when the input sequence is provided, the system would be able to generate the correct sequences from that root word. This would reduce the size of the dictionary as similar derivational words will be removed (eg. 'पुस्तक', 'पुस्तकों', 'पुस्तके' can all be captured using the same root word 'पुस्तक'). Further we would like to improve our system to provide more optimized predictions and adaptability of user and evaluation of linguistic contexts as discussed by Hasselgren et al [7]. We are under process of developing a system which can readjust itself and predict user's natural preference of input, thus making the system more user friendly.